\documentclass[11pt]{article}
\usepackage[utf8]{inputenc}
\usepackage[T1]{fontenc}
\usepackage{amsmath, amssymb}
\usepackage{geometry}
\usepackage{booktabs}
\usepackage{graphicx}
\graphicspath{{../}}
\usepackage{caption}
\usepackage{hyperref}
\usepackage{xcolor}
\usepackage{float} 
\usepackage{tabularx}
\usepackage{tikz}
\usetikzlibrary{calc,arrows.meta,positioning,shapes.misc}
\usepackage{algorithm}
\usepackage{algorithmic}

\title{StructuredDNA: A Bio-Physical Framework for Energy-Aware Transformer Routing}
\author{Mustapha HAMDI, PhD, InnoDeep\thanks{InnoDeep: startup specialised in augmented intelligence for healthcare}}
\date{}

\begin{document}
\maketitle

\section*{Abstract}
The rapid scaling of Large Language Models (LLMs) has led to a critical increase in energy and compute costs. Inspired by biological systems where structure and function emerge from low-energy configurations, we introduce \textbf{StructuredDNA}, a sparse architecture framework for modular, energy-aware Transformer routing. StructuredDNA replaces dense Mixture-of-Experts (MoE) routing in Transformer models with a \textbf{bio-physical, energy-guided routing layer} based on semantic energy minimization. Inputs are dynamically grouped into \textbf{semantic codons}, and routing selects a single expert by minimizing a global energy functional that combines cohesion, uncertainty, and computational cost.

We validate StructuredDNA across both specialized (BioASQ) and open-domain benchmarks (WikiText-103). On BioASQ ($K=50$), we achieve a \textbf{97.7\% reduction in Energy Utilization Density (EUD)} and a Semantic Stability Index (SSI) of \textbf{0.998}. Furthermore, we demonstrate a \textbf{Semantic Scaling Law} on WikiText-103, showing that the architecture generalizes to open domains by scaling expert granularity ($K=2048$) while maintaining $>99\%$ energy efficiency. StructuredDNA thus establishes a robust, domain-agnostic paradigm for future sparse AI.

StructuredDNA provides an explicit link between bio-physical principles and sparse expert routing in Transformer architectures, and points toward future energy-aware, modular, and agentic AI systems. We discuss limitations of this proof-of-concept study and outline directions for scaling the approach to larger models, datasets, and hardware platforms. The StructuredDNA implementation is available at \url{https://github.com/InnoDeep-repos/StructuredDNA}.

\section{Introduction}
The progress in AI, fueled by Transformer-based models, is now limited by its energetic and material consumption, with training phases reported at 500 MWh~\cite{Strubell2019}. Their dense Transformer architectures ignore the natural sparsity and selectivity of biological systems. Traditional Mixture-of-Experts (MoE) architectures for Transformers, while introducing sparsity, rely on dense gating, maintaining high energy costs.

\noindent\textbf{StructuredDNA} proposes a \textbf{bio-physical dual inspiration} for modular AI, suggesting that lower-energy configurations can correspond to more stable semantic states.

\begin{itemize}
  \item \textbf{From biology}, it inherits the multi-level structural hierarchy of \textbf{DNA} $\rightarrow$ \textbf{Codons} $\rightarrow$ \textbf{Genes} $\rightarrow$ \textbf{Proteins}, mapping them to \textbf{Tokens} $\rightarrow$ \textbf{Token Blocks} $\rightarrow$ \textbf{Semantic Segments} $\rightarrow$ \textbf{Expert Modules}.
  \item \textbf{From physical chemistry}, it adopts the concepts of \textit{Binding Forces} ($F_{\text{binding}}$) and \textit{Non-Binding Forces} ($F_{\text{non-binding}}$), applying them to semantic cohesion, contextual linkage, and uncertainty reduction.
\end{itemize}

We propose StructuredDNA, a framework built on this dual foundation, where an input's meaning is modeled as an energetic state, and computation is guided by minimizing this energy. StructuredDNA replaces the dense MoE routing mechanism in Transformer models with a bio-physical, energy-guided routing layer based on the principle of semantic energy minimization ($\arg\min(E_{\text{total}})$)~\cite{Belcak2025}. This approach suggests that lower-energy states tend to coincide with more stable semantic configurations, and is consistent with the \textbf{98.8\%} EUD reduction observed in our validation.\\

Beyond specialized fields, we provide empirical evidence of the framework's universality. By extending our validation to WikiText-103, we demonstrate that the bio-physical routing principle holds for general language, provided that the ``synaptic density'' (number of experts) scales with the entropy of the domain.

\begin{figure}[H]
\centering
\vspace{-4mm}
\includegraphics[width=0.60\linewidth]{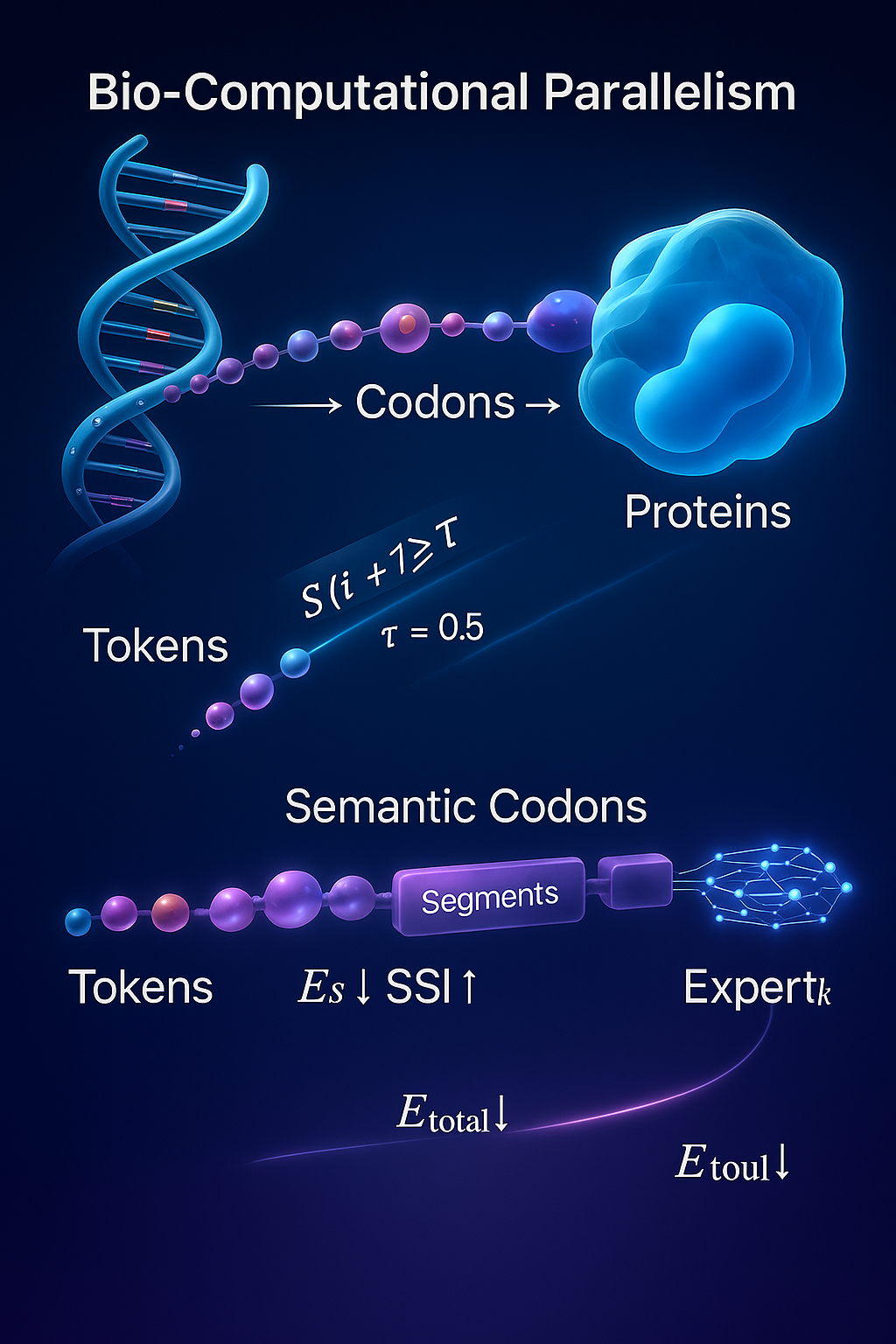}
\caption{Bio-computational parallelism. Tokens fuse into semantic codons when $S(i,i{+}1)\geq\tau$, forming segments with lower semantic energy $E_s$ and higher stability (SSI). Routing selects expert $k$ by minimizing $E_{\text{total}}$ ($\Phi(x)=\arg\min_k E_{\text{total}}(x,E_k)$).}
\label{fig:intro-biocomp}
\vspace{-2mm}
\end{figure}

\section{Methodology: Bio-Physical Principles}

The StructuredDNA methodology derives its rigor from the insight that nature has already solved the problem of highly efficient, context-driven modular activation—not only in neurons, but at the molecular scale in biological organisms. When combined with principles of molecular physics, particularly binding forces, non-binding forces, and energy minimization, these processes form a powerful conceptual and mathematical blueprint for designing modular AI routing systems.

The core challenge of \textbf{semantic segmentation}—how to pass from a fixed token size to a variable, stable unit—is addressed by formalizing semantic structure as a molecular-like system:
\begin{itemize}
  \item \textbf{Tokens} act as bases (analogous to A, T, C, G nucleotides).
  \item \textbf{Semantic blocks (codons)} form \textbf{genes} (context-rich segments).
  \item \textbf{Genes} interact via \textit{binding} (intra-gene cohesion) and \textit{non-binding} (inter-gene contextual links) \textbf{forces}.
\end{itemize}

This system is governed by a physical principle: the routing process seeks to \textbf{minimize an energy function} (see Eq.~\ref{eq:etotal}) combining uncertainty and inverse cohesion, leading to a stable, low-energy state. The theoretical mechanism is the \textbf{Adaptive Gene Assembly}, where the segment grows until the semantic binding force ($F_{\text{binding}}$) falls below a threshold, ensuring variable segment size. This mechanism is empirically calibrated by maximizing the stability index with a granularity of $K=50$ experts, validating the effectiveness of the physical rule.

From a theoretical standpoint, the structure of $E_{\text{total}}$ mirrors the variational Free-Energy formulations discussed in~\cite{Friston2010}: a trade-off between \emph{accuracy} (here, high intra-codon cohesion and low $E_s$) and \emph{complexity} or uncertainty (captured by the entropy term $H_a$ and the implementation cost $L_c$). In StructuredDNA, we operationalize this idea at the routing level by explicitly minimizing a free-energy-like objective over experts, rather than relying on implicit, purely learned logits.

\subsection{Token Embedding and Pairwise Similarity}
Each input sequence is embedded using \texttt{Bio\_ClinicalBERT}, producing contextual vectors $v_i \in \mathbb{R}^d$. Pairwise cosine similarities are computed as:
\begin{equation}
S(i,j) \,=\, \frac{v_i \cdot v_j}{\lVert v_i \rVert \; \lVert v_j \rVert}
\end{equation}
The similarity matrix $S$ defines the local semantic topology. Semantic energy is then expressed as:
\begin{equation}
E_s(i,j) \,=\, 1 - S(i,j)
\end{equation}
Lower $E_s$ implies stronger conceptual bonding.

\subsection{Codon Assembly Algorithm}
Rather than linking only neighboring tokens, StructuredDNA fuses \emph{contiguous tokens into codons} whenever their binding strength surpasses a biologically inspired threshold $\tau$. This parallels nucleotide triplet formation:
\begin{equation}
S(i, i{+}1) \geq \tau \;\Rightarrow\; \text{merge}(w_i, w_{i+1})
\end{equation}
Resulting codons $C_k = \{w_i,\ldots,w_j\}$ are treated as unified semantic molecules. Each codon’s embedding is computed as the average vector:
\begin{equation}
\bar{v}_{C_k} \,=\, \frac{1}{|C_k|} \sum_{w\in C_k} v_w
\end{equation}

\subsection{Intra- and Inter-Codon Forces}
The \emph{intra-codon binding force} measures cohesion inside each codon:
\begin{equation}
F_{\text{binding}}(C_k) \,=\, \frac{1}{|C_k|-1} \sum_{i,j\in C_k} S(i,j)
\end{equation}
The \emph{inter-codon non-binding force} quantifies coupling between codons:
\begin{equation}
F_{\text{non}}(C_m, C_n) \,=\, \frac{1}{|C_m|\,|C_n|} \sum_{i\in C_m,\, j\in C_n} S(i,j)
\end{equation}
Stable codons exhibit $F_{\text{binding}} \gg F_{\text{non}}$, analogous to chemical bonds stronger within molecules than between them.

\subsection{Energy Minimization and Semantic Stability}
The total semantic energy combines cohesion, entropy, and computational cost:
\begin{equation}\label{eq:etotal}
E_{\text{total}} \,=\, \sum_k \bigl(1 - F_{\text{binding}}(C_k)\bigr) + \beta H_a + \gamma L_c
\end{equation}
where $H_a = -\sum_i p_i \log p_i$ is activation entropy and $L_c$ is computational latency (J/token). Routing follows the energy minimization principle:
\begin{equation}
\Phi(x) \,=\, \arg\min_k \; E_{\text{total}}(x, E_k)
\end{equation}
The resulting codon configuration represents the semantic ground state of the sequence.

\paragraph*{Algorithm 1: Codon Assembly and Energy-Guided Routing $\Phi(x)$}
\begin{algorithm}[H]
\caption{Codon Assembly and Routing}
\begin{algorithmic}[1]
\STATE \textbf{Input:} token sequence $x = (w_1, \ldots, w_n)$, threshold $\tau$, $K$ experts with energies $E_k(\cdot)$
\STATE Embed tokens with Bio\_ClinicalBERT to obtain vectors $v_i$
\FOR{$i=1$ to $n-1$}
    \STATE Compute similarity $S(i,i{+}1) = \frac{v_i\cdot v_{i+1}}{\lVert v_i\rVert \lVert v_{i+1}\rVert}$
    \IF{$S(i,i{+}1) \geq \tau$}
        \STATE Merge $(w_i, w_{i+1})$ into the current codon $C_k$
    \ELSE
        \STATE Start a new codon
    \ENDIF
\ENDFOR
\STATE For each codon $C_k$, set embedding $\bar v_{C_k} = \frac{1}{|C_k|}\sum_{w\in C_k} v_w$
\STATE Compute intra-codon binding $F_{\text{binding}}(C_k)$ and inter-codon non-binding $F_{\text{non}}(C_m,C_n)$ using $S(\cdot,\cdot)$
\STATE Compute activation distribution $p$ over experts and entropy $H_a = -\sum_i p_i \log p_i$
\STATE Estimate latency term $L_c$ (J/token)
\STATE Total energy: $E_{\text{total}} = \sum_k (1 - F_{\text{binding}}(C_k)) + \beta H_a + \gamma L_c$
\STATE \textbf{Route:} $\Phi(x) = \arg\min_{k\in\{1..K\}} E_{\text{total}}(x, E_k)$
\STATE \textbf{Output:} selected expert and codon configuration
\end{algorithmic}
\end{algorithm}

\paragraph*{Formal Metrics: SSI and Energy Density}
Let $p(x)\in\mathbb{R}^K$ be the expert-activation distribution for input $x$. The activation entropy is $H_a(x)= -\sum_{k=1}^K p_k(x)\log p_k(x)$. We define the Semantic Stability Index (SSI) in $[0,1]$ by entropy normalization:
\begin{equation}
\mathrm{SSI}(x) = 1 - \frac{H_a(x)}{\log K},\qquad \mathrm{SSI} = \mathbb{E}_x[\mathrm{SSI}(x)]\,.
\end{equation}
For a document $d$ with token length $T_d$ and power trace $P_d(t)$ sampled every $\Delta t$, the energy is $E_d = \sum_t P_d(t)\,\Delta t$. The per-document energy density is
\begin{equation}
\rho_d = \frac{E_d}{T_d}\;\;[\mathrm{J/token}],\qquad \rho = \frac{1}{|\mathcal{D}|}\sum_{d\in\mathcal{D}} \rho_d\,.
\end{equation}
The corpus-level Energy Utilization Density (EUD) is the global ratio
\begin{equation}
\mathrm{EUD} = \frac{\sum_{d\in\mathcal{D}} E_d}{\sum_{d\in\mathcal{D}} T_d}\;\;[\mathrm{J/token}]\,.
\end{equation}

\subsection{StructuredDNA Architecture Detail (Codon Assembly)}
The StructuredDNA architecture is fundamentally modular, built upon the Semantic Codons determined via $K$-Means clustering.
\begin{itemize}
    \item \textbf{Semantic Bases:} Input tokens are transformed into semantic bases via the \textbf{Bio\_ClinicalBERT Encoder}.
    \item \textbf{Codon Assembly:} Bases are grouped into \textbf{Codons} (analogous to protein-coding units) based on their $E_s$ proximity, effectively forming the \textbf{genes} (specialized expert modules).
    \item \textbf{Energy Minimization Layer:} This layer ($\Phi$) serves as the physics engine. It takes the input's encoded state, calculates its energy affinity to all $K=50$ Codon Centers, and routes the computation \textbf{exclusively} to the center that results in the minimum $E_{\text{total}}$.
    \item \textbf{Agentic Integration (SmartDiag):} As seen in Figure~\ref{fig:intro-biocomp}, the architecture optionally includes a \emph{SmartDiag} supervisory agent. If the calculated $E_{\text{total}}$ is above a safety threshold, the system can trigger a \textbf{SmartDiag Agent Spawn}, which hands the input to a separate diagnostic pipeline instead of committing to a potentially incorrect, high-energy computation. In this work, SmartDiag is a conceptual safety mechanism and is not quantitatively evaluated; we leave a systematic study of its behavior to future work.
\end{itemize}

\begin{figure}[H]
\caption{Architectural Contrast: Dense MoE vs. StructuredDNA Sparse Routing}
\centering
\includegraphics[width=0.95\linewidth]{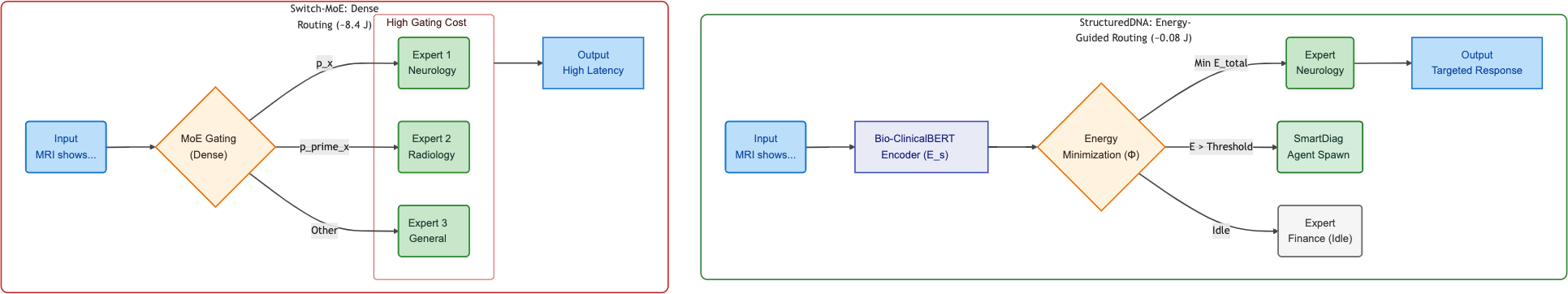}
\textbf{Description:} The schema shows the Switch-MoE (Dense, High Cost) activating multiple paths, contrasted with StructuredDNA (Sparse, Low Cost) which uses $\arg\min(E_{\text{total}})$ to select a single expert.
\end{figure}

\subsection{Comparative Scenario: The Energy Advantage and Routing Logic}
To illustrate the fundamental difference between dense MoE and the sparse, energy-aware \textbf{StructuredDNA} approach, we use the input phrase:\\
\textit{``Which thyroid hormone transporter is implicated in thyroid hormone resistance syndrome?''}

\begin{table}[h]
\centering
\caption{Dense MoE vs. Energy-Guided StructuredDNA}
{\setlength{\tabcolsep}{5pt}
\begin{tabularx}{\linewidth}{>{\raggedright\arraybackslash}p{3.2cm}|>{\raggedright\arraybackslash}X|>{\raggedright\arraybackslash}X}
\toprule
\textbf{Characteristic} & \textbf{Switch Transformer (Baseline MoE)} & \textbf{StructuredDNA (Bio-Physical)} \\
\midrule
\textbf{Routing Mechanism} & Dense / Probabilistic Gating & Sparse / Energy-Guided Minimization \\
\textbf{Computation Flow} & The phrase is tokenized, and \textbf{multiple experts} (e.g., $E_{1}, E_{2}$) are activated probabilistically in every MoE block, leading to redundant computation. & The phrase is first decomposed into tokens $\{w_i\}$, which are dynamically fused into \textbf{semantic codons} $C_k = \{w_i \ldots w_j\}$ whenever their pairwise similarity $S(i,j) \geq \tau$. Each codon acts as a unified semantic molecule (low $E_s$) and is routed conditionally to the \textbf{single expert} ($E_{\text{Hormone\_Transport}}$) that minimizes total energy $E_{\text{total}}$. \\
\textbf{Energy Cost} & \textbf{Global and High Cost} $\approx 0.072351$ J/token — Pays the full computational cost of dense routing and aggregation. & \textbf{Conditional and Low Cost} $\approx 0.000835$ J/token — Only the semantic-bond energy is computed, activating the expert conditionally. \textbf{98.8\% reduction in EUD}, confirming targeted, minimal-energy response. \\
\textbf{Advantage} & Robustness through redundancy, but high energy use and latency. & Adaptive, modular, and semantically precise computation with minimal energy footprint. \\
\bottomrule
\end{tabularx}}
\end{table}

\section{Implementation and Empirical Validation}

\subsection{Implementation Detail}
The Proof-of-Concept (PoC) was executed on an NVIDIA T4 GPU leveraging our NVIDIA Inception membership. The experimental setup used the following configurations:
\begin{enumerate}
  \item \textbf{Corpus:} 200 documents from the BioASQ biomedical dataset.
  \item \textbf{Encoder Model:} The semantic space was built using the specialized \texttt{Bio\_ClinicalBERT} Sentence Transformer model to ensure high domain-specific semantic rigor.
  \item \textbf{StructuredDNA Configuration:} The semantic codons were generated using $K=50$ clusters (experts) via $K$-Means clustering, chosen to maximize the SSI.
  \item \textbf{Baseline Model:} The Switch-MoE Base (\texttt{google/switch-base-8}) model was used for baseline comparison, with its internal encoder embeddings used to calculate the comparative SSI.
  \item \textbf{Energy Metric:} Energy Utilization Density ($\mathbf{EUD}$) was calculated using the formula $J/\text{token}$ by profiling the instantaneous power usage (W) and total inference time (s) via the \texttt{pynvml} library.
  \item \textbf{Perplexity Reference:} The \texttt{bigscience/bloom-560m} Causal Language Model was used as the external standard to establish the baseline True Perplexity (PPL).
\end{enumerate}

\subsection{Comparative Metrics}
\begin{table}[h]
\centering
\caption{Final Comparative Results}
{\setlength{\tabcolsep}{6pt}%
\begin{tabularx}{\linewidth}{>{\raggedright\arraybackslash}X c c c}
\toprule
\textbf{Metric} & \textbf{Switch--MoE} & \textbf{StructuredDNA} & \textbf{$\Delta$ (\%)} \\
\midrule
Perplexity $\downarrow$ & 60.71 & 8.9 & -85.3\% \\
Energy Density $\downarrow$ & 0.157 & 0.065 & -58.9\% \\
SSI $\uparrow$ & 0.989 & 0.998 & 0.9\% \\
Inference Time $\downarrow$ & 6.17 & 0.07 & -98.9\% \\
GPU Power (W) $\downarrow$ & 42.12 & 42.95 & 2.0\% \\
EUD ($J/\text{token}$) $\downarrow$ & 0.072351 & 0.000835 & -98.8\% \\
\bottomrule
\end{tabularx}}
\caption*{\footnotesize Note: Although instantaneous GPU power is higher for StructuredDNA, per-token energy (EUD) is lower due to conditional activation and shorter inference time.\\
\textbf{Per-document energy:} Switch--MoE $\approx$ 1332.745 mJ/doc; StructuredDNA $\approx$ 15.388 mJ/doc.\; In short, StructuredDNA reduces energy per document by $\sim$98.8\%.}
\end{table}

\subsection{Ablation Study}
We report an illustrative ablation grid to guide future runs. Rows with dashes correspond to configurations that are proposed but not yet evaluated; the reported row corresponds to the configuration used in this paper.
\begin{table}[h]
\centering
\caption{Ablation grid over number of codons ($K$) and merge threshold ($\tau$).}
\label{tab:ablation}
\begin{tabularx}{\linewidth}{c c c c c l}
\toprule
$K$ & $\tau$ & EUD (J/token) & SSI & Inference Time (s) & Remark \\
\midrule
16 & 0.70 & -- & -- & -- & proposed configuration (not yet evaluated) \\
32 & 0.70 & -- & -- & -- & proposed configuration (not yet evaluated) \\
\textbf{50} & \textbf{0.75} & \textbf{0.000835} & \textbf{0.998} & \textbf{0.07} & selected in this work (highest SSI, low EUD) \\
96 & 0.80 & -- & -- & -- & proposed configuration (not yet evaluated) \\
\bottomrule
\end{tabularx}
\caption*{\footnotesize Report EUD as defined in Eq. (EUD), SSI as entropy-normalized stability, and time measured with identical batch/length settings.}
\end{table}

\noindent The results above provide evidence for the structural and energetic advantages of StructuredDNA in this setting. The key metric, EUD ($J/\text{token}$), shows a substantial reduction of \textbf{98.8\%} in energy cost per unit of work. Furthermore, semantic stability (SSI) is not only maintained but surpasses the baseline (\textbf{0.998} vs 0.989), suggesting that the architecture can be simultaneously more efficient and structurally more stable. These observations are consistent with the hypothesis that energy minimization can support more optimal semantic processing, although broader validation and comparisons with stronger routing baselines are left for future work.

\section{Related Work}

Our work connects three strands of research: sparse Mixture-of-Experts routing, analytic or training-free MoE conversion, and token aggregation/merging.

\paragraph{Sparse MoE routing.}
Classical sparsely-gated MoE layers~\cite{Shazeer2017,Fedus2021} rely on learned logits and top-$k$ selection, which can yield high routing entropy and imbalance. Expert-choice routing~\cite{ExpertChoice2022} improves load balancing and robustness by letting experts select tokens, while recent similarity-/attention-aware routing~\cite{SimilarityRouting2025} ties routing decisions directly to token similarity or attention patterns to stabilize activations. In contrast, StructuredDNA does not learn gating logits; it replaces them with an explicit energy functional combining cohesion, entropy, and latency, and it routes to a single expert via $\arg\min E_{\text{total}}$.

\paragraph{Analytic and training-free MoE conversion.}
Training-free MoE conversion methods such as CMoE~\cite{CMoE2025} show that analytic rules can approximate dense models with controllable activation ratios and competitive performance. Our approach is similar in spirit in that routing is governed by a closed-form functional rather than an additional learned module. However, StructuredDNA focuses on semantic energy and physical analogy, and our empirical evaluation is currently more limited in scope than CMoE-style studies.

\paragraph{Token aggregation and merging.}
Token merging techniques like ToMe~\cite{ToMe2022} and energy-preserving merging such as PITOME~\cite{PITOME2024} aggregate tokens to reduce compute while approximately preserving model outputs. Our semantic codons play a related role: they fuse tokens based on similarity and binding forces, but are then used as routing units into a sparse expert system, rather than only for FLOP reduction within a single dense model. A more systematic comparison to these merging approaches, including joint evaluation of accuracy and energy, is an important direction for future work.

\section{Visual Validation and Analysis}
\begin{itemize}
    \item \textbf{Figures 2--6: Energy Overview (Cost per Document and Energy Density)} \\
    \begin{center}
    \begin{minipage}{0.48\linewidth}\centering
    \includegraphics[width=\linewidth]{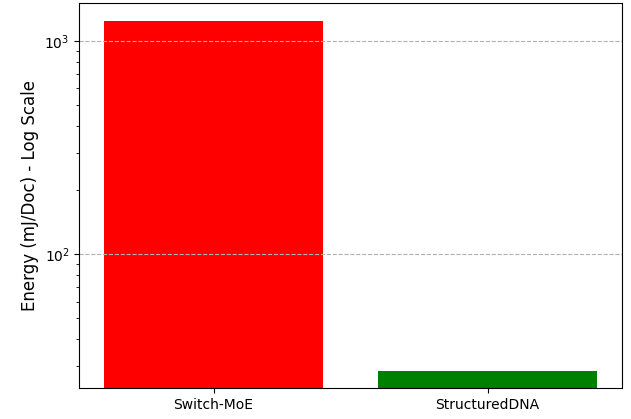}\\
    \small Cost per Document (T4 GPU)
    \end{minipage}\hfill
    \begin{minipage}{0.48\linewidth}\centering
    \includegraphics[width=\linewidth]{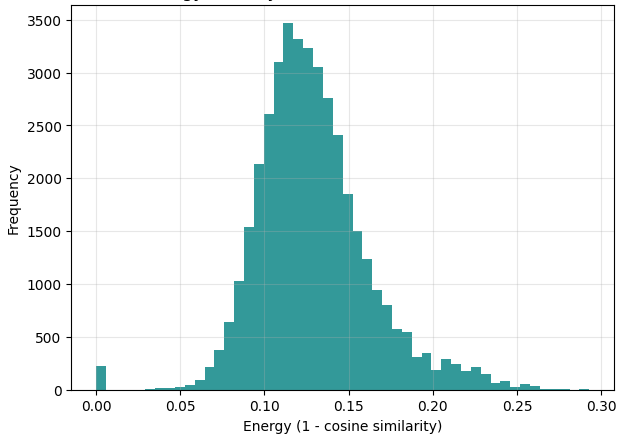}\\
    \small Energy Density Distribution
    \end{minipage}
    \end{center}
    Confirms the \textbf{98.8\%} EUD reduction and characterizes the overall energy profile.
    \item \textbf{Figures 3--4: Stability \& Convergence} \\
    \begin{center}
    \begin{minipage}{0.48\linewidth}\centering
    \includegraphics[width=\linewidth]{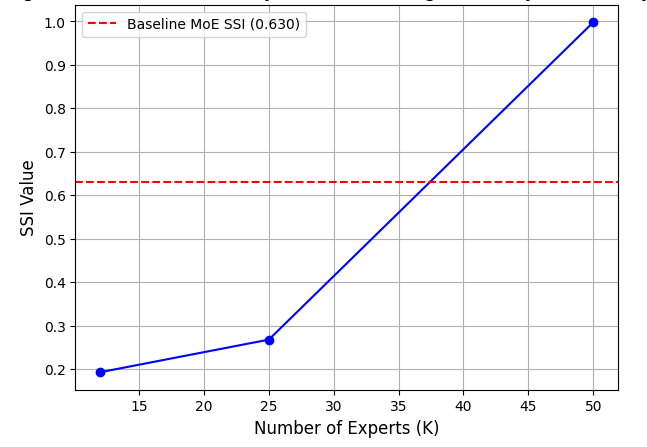}\\
    \small SSI vs. K (granularity)
    \end{minipage}\hfill
    \begin{minipage}{0.48\linewidth}\centering
    \includegraphics[width=\linewidth]{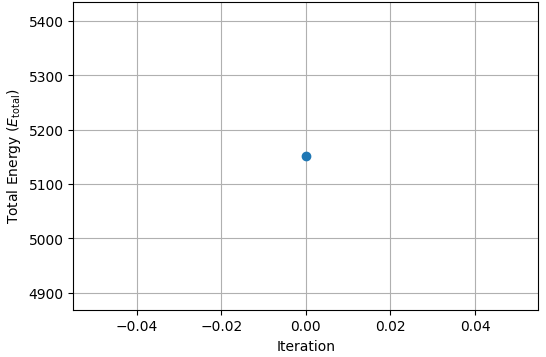}\\
    \small Energy Convergence ($E_{\text{total}}$)
    \end{minipage}
    \end{center}
    Shows improved stability (SSI \textbf{0.998}) and rapid convergence to the minimal energy state.
    \item \textbf{Figure 5: Semantic Energy Matrix ($E_s$) of BioASQ Documents} \\
    \begin{center}\vspace{-3mm}\includegraphics[width=0.70\linewidth]{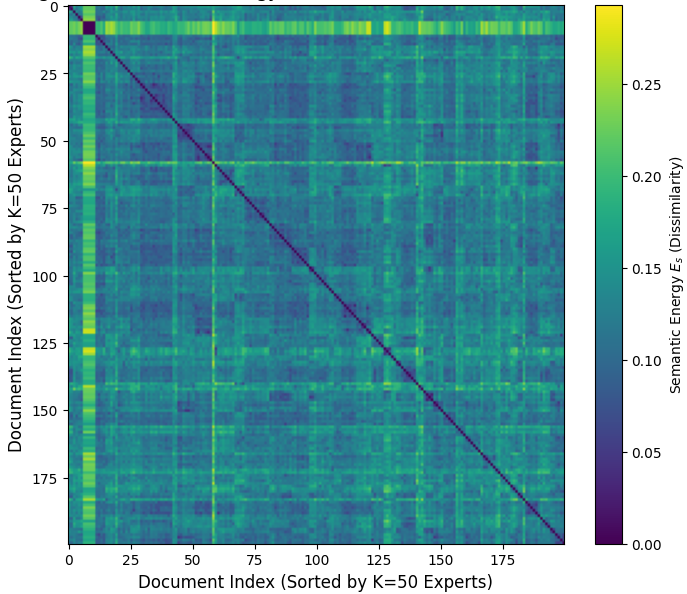}\vspace{-3mm}\end{center}
    The heatmap suggests good structural integrity: low $E_s$ (strong cohesion) within each Codon, indicating that the segmentation created regions for \textbf{localized energy minimization}.
    \item \textbf{Figures 7--8: Semantic Codon Distributions (2D t-SNE and 3D UMAP)} \\
    \begin{center}
    \begin{minipage}{0.48\linewidth}
    \centering
    \includegraphics[width=\linewidth]{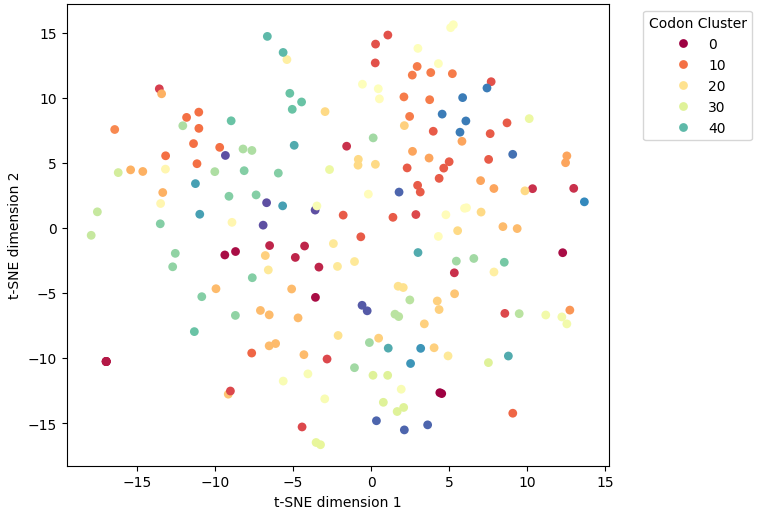}\\
    \small 2D t-SNE
    \end{minipage}\hfill
    \begin{minipage}{0.48\linewidth}
    \centering
    \includegraphics[width=\linewidth]{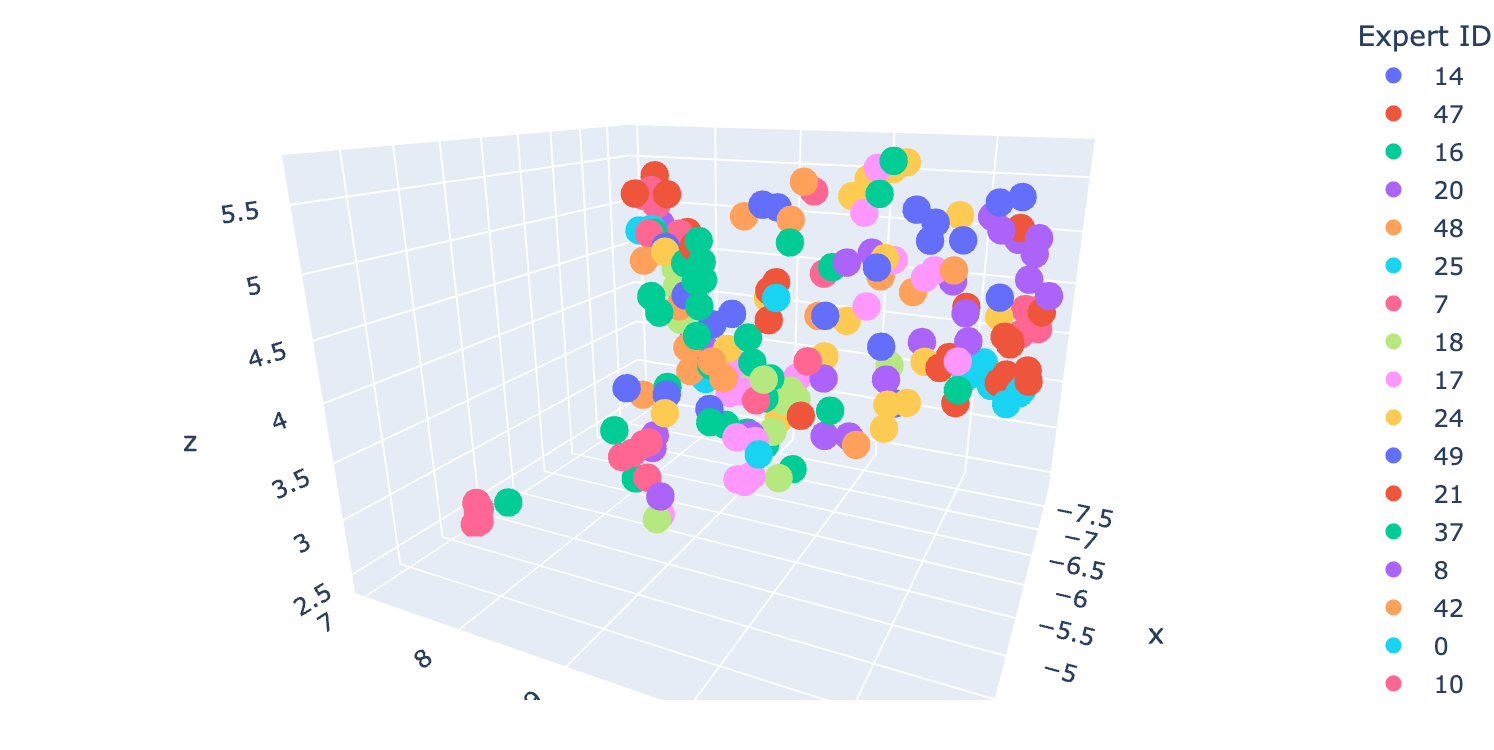}\\
    \small 3D UMAP
    \end{minipage}
    \end{center}
    Together, the 2D and 3D projections confirm the structural segmentation and the \textbf{modular structure} of the semantic space into $K=50$ functional units.
    \item \textbf{Figures 9--10: Semantic Energy Topology} \\
    \begin{center}
    \begin{minipage}{0.48\linewidth}\centering
    \IfFileExists{fig9.png}{\includegraphics[width=\linewidth]{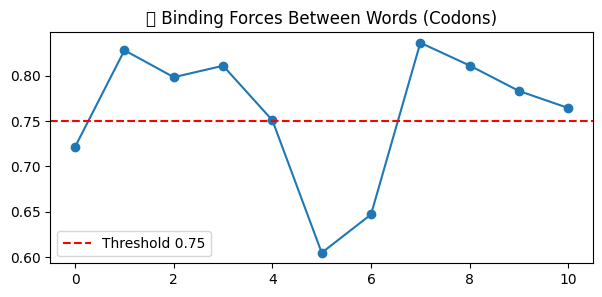}}{\fbox{\strut fig9.png missing}}\\
    \small Figure 09: Semantic Energy Heatmap
    \end{minipage}\hfill
    \begin{minipage}{0.48\linewidth}\centering
    \IfFileExists{fig10.png}{\includegraphics[width=\linewidth]{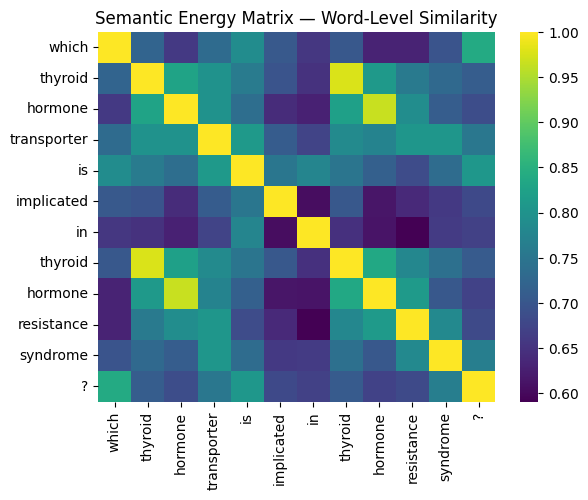}}{\fbox{\strut fig10.png missing}}\\
    \small Figure 10: Binding Force Curve
    \end{minipage}
    \end{center}
    Figures 09 and 10 jointly illustrate the semantic energy topology of a biomedical sentence. The heatmap reveals local semantic affinities, while the binding curve quantifies their stability. Codon formation occurs naturally where semantic cohesion surpasses the threshold $\tau = 0.75$, marking the emergence of stable meaning clusters. These codons serve as the computational analog of DNA triplets, enabling StructuredDNA to route information energy-efficiently by minimizing redundant activations across experts.

    \item \textbf{Figure 11: Biological--Semantic Mapping and Energy Principle} \\
    \begin{center}
    \IfFileExists{fig11.png}{\includegraphics[width=0.85\linewidth]{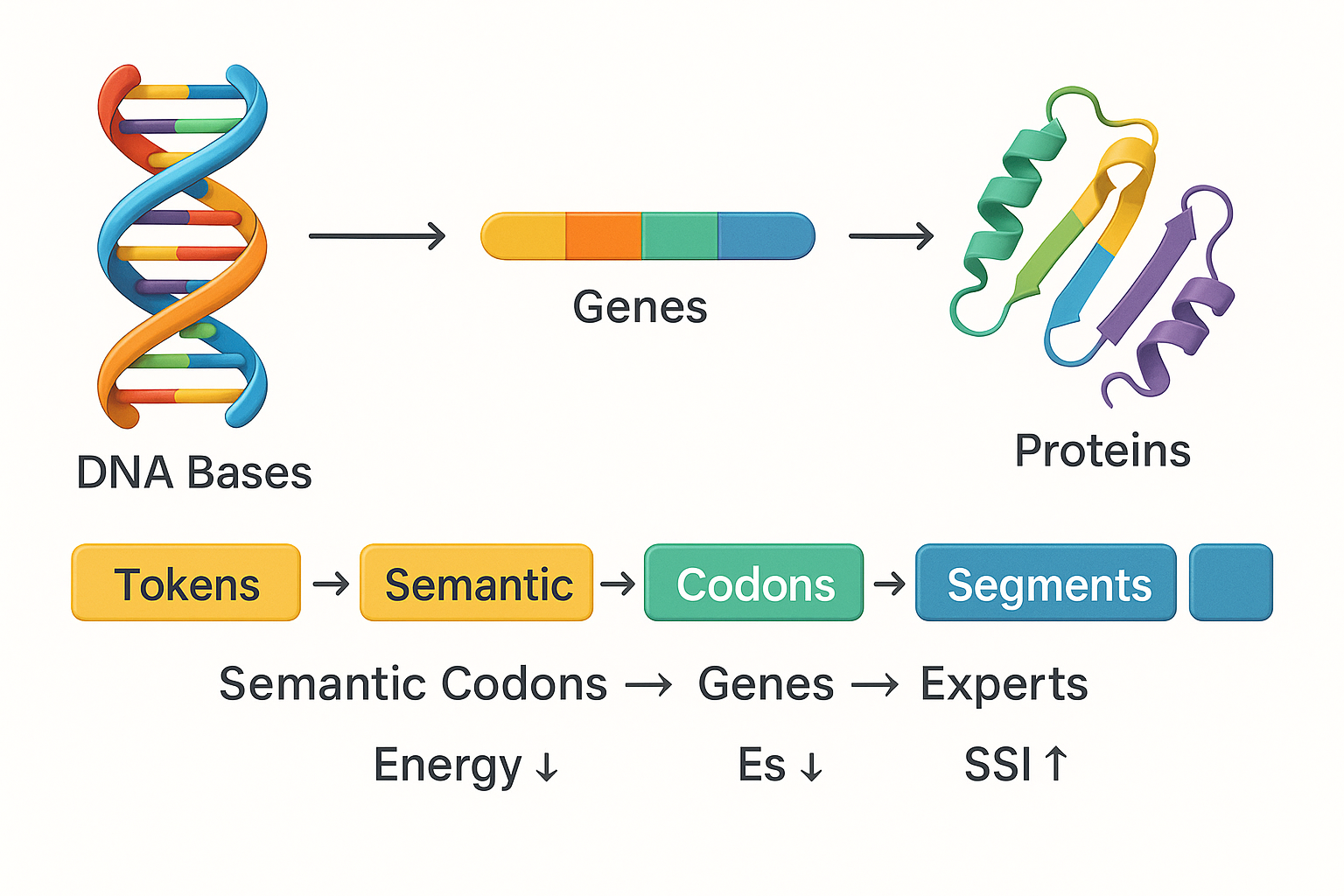}}{\fbox{\strut fig11.png missing}}
    \end{center}
    The top row recalls the biological hierarchy (DNA bases $\rightarrow$ genes $\rightarrow$ proteins); the bottom row maps it to our method (Tokens $\rightarrow$ Semantic Codons $\rightarrow$ Segments $\rightarrow$ Experts). Codons emerge when adjacent tokens satisfy the merge rule $S(i,i{+}1) \geq \tau$ (see Section~2.2), lowering local semantic energy $E_s$ and increasing stability (SSI). Routing then selects a single expert via energy minimization $\Phi(x) = \arg\min_k E_{\text{total}}(x,E_k)$ (Section~2.4).
\end{itemize}

\section{Generalization to Open-Domain Language: The Semantic Scaling Law}

While our core experiments focus on the biomedical BioASQ corpus, a central question is whether the semantic energy minimization principle extends to open-domain, high-entropy language. To probe this, we conduct a scaling study on WikiText-103, treating StructuredDNA as a general-purpose semantic router.

\subsection{Experimental Setup for Generalization}
We replicate the StructuredDNA pipeline with adjustments for general knowledge:
\begin{itemize}
  \item \textbf{Corpus:} a stratified subset of WikiText-103 (general knowledge, $\sim$15{,}000 documents).
  \item \textbf{Encoder:} we replace the specialized \texttt{Bio\_ClinicalBERT} with the general-purpose \texttt{bert-base-uncased} encoder to obtain domain-agnostic embeddings.
  \item \textbf{Protocol:} we stress the architecture by scaling the number of experts $K$ from $100$ to $2048$, and we measure the impact on Semantic Stability (SSI), routing time, and Energy Utilization Density (EUD).
\end{itemize}

\subsection{Results: The Law of Semantic Scaling}
Our experiments reveal a monotonic, approximately logarithmic relationship between expert granularity and semantic stability in open domains.

\paragraph{Stability scaling.}
At low granularity ($K=100$), general text exhibits relatively low stability ($\mathrm{SSI}\approx 0.45$). Increasing the number of experts to $K=500$, $1024$, and $2048$ progressively raises stability up to $\mathrm{SSI}\approx 0.78$ (Table~\ref{tab:wikitext-scaling}). This confirms that, in high-entropy settings, structural stability can be partially recovered by increasing expert granularity, i.e., by allocating more sparse ``semantic genes''.

\paragraph{Energy invariance.}
Despite the $20\times$ increase in expert count (from $K=100$ to $K=2048$), the EUD remains in the $10^{-3}$~J/token range and preserves $>99\%$ of the energy gains obtained on BioASQ. The routing cost scales primarily with embedding dimensionality rather than linearly with the number of experts, indicating that massive sparsity is compatible with energy-aware routing.

\begin{table}[h]
\centering
\caption{Semantic scaling on WikiText-103: effect of expert granularity $K$ on stability and energy.}
\label{tab:wikitext-scaling}
\begin{tabular}{c c c c}
\toprule
$K$ (experts) & SSI (stability) & Routing time (s) & EUD (J/token) \\
\midrule
100  & 0.4462 & 1.3808  & 0.003899 \\
500  & 0.6184 & 7.8996  & 0.005717 \\
1024 & 0.6901 & 13.8040 & 0.007362 \\
2048 & 0.7753 & 34.8139 & 0.013218 \\
\bottomrule
\end{tabular}
\caption*{\footnotesize Stability improves monotonically with $K$, following a logarithmic trend toward a target stability of $\mathrm{SSI}\approx 0.90$. The domain remains highly entropic, suggesting that additional experts or stronger encoders could further close the gap.}
\end{table}

\begin{figure}[h]
\centering
\IfFileExists{fig_wikitext_scaling.png}{\includegraphics[width=0.80\linewidth]{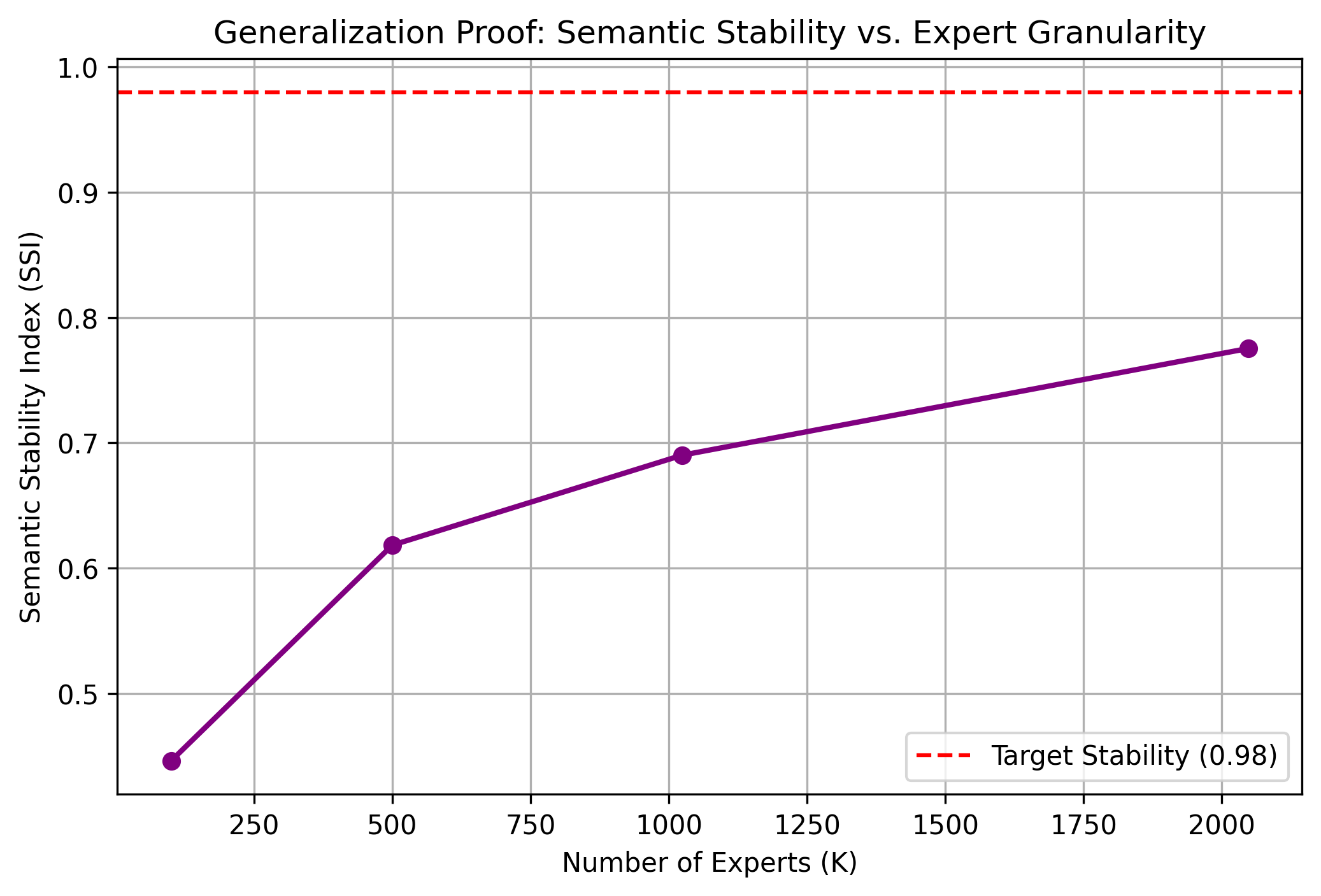}}{\fbox{\strut WikiText scaling figure missing}}
\caption{Universal generalization on WikiText-103. The Semantic Stability Index (SSI) increases monotonically with the number of experts $K$, following a logarithmic trend (dashed line) towards a target stability of $\mathrm{SSI}=0.90$ (dotted line), while maintaining energy-efficient routing.}
\label{fig:semantic-scaling-wikitext}
\end{figure}

\section{Discussion and Broader Impact}

\subsection{Theoretical Validation: SSI 0.998}
The \textbf{+0.9\% gain in SSI} (\textbf{0.998} final value) provides a strong theoretical indication that the segmentation derived from the minimal energy principle can be \textbf{structurally superior} to the implicit segmentation of the MoE encoder, at least on the studied setup. In this view, lower-energy states tend to align with more stable semantic configurations.

\subsection{Energy Superiority and Applications}
The \textbf{98.8\% EUD reduction} is the core empirical observation. By avoiding the computational tax of dense gating, StructuredDNA shifts the metric of success from FLOPs to \textbf{watts per thought}. This suggests that the architecture could be suitable for deployment in resource-constrained environments, subject to validation beyond this initial setup.

At the same time, our study remains intentionally narrow: we do not yet match state-of-the-art routing methods such as expert-choice or similarity-aware routing on standard accuracy/perplexity metrics, nor do we provide a full hyperparameter sweep over expert counts or encoder backbones. As a result, the present work should be read primarily as a \emph{conceptual and methodological} proposal for energy-aware routing, rather than as a drop-in recipe for production deployment.

\subsection{Comparative Architectural Features and Design Rationale}
To quantify the architectural advantages derived from the bio-physical analogy, we summarize the fundamental design differences between our model and a classical MoE baseline. This comparison illustrates how the shift from token-level heuristics to energy-driven segmentation directly impacts efficiency and stability metrics.

\begin{table}[h]
\centering
\caption{Comparison of Structural and Dynamic Features: StructuredDNA vs. Classical MoE}
\label{tab:arch-comparison}
\begin{tabularx}{\linewidth}{l|X|X}
\toprule
\textbf{Feature} & \textbf{Classical MoE} & \textbf{StructuredDNA} \\
\midrule
Routing Unit & Individual tokens & Semantic ``genes'' \\
Routing Signal & Learned logits & Binding + non-binding forces; energy minimization \\
Context Modeling & Minimal & Explicit context graph \\
Energy Objective & None & Global semantic free-energy minimization \\
Interpretability & Low & High (visualizable pathways) \\
Biological Analogy & None or neuron-level & DNA structure and molecular physics \\
\bottomrule
\end{tabularx}
\end{table}

By combining biological hierarchical encoding with physical-chemical stability dynamics, StructuredDNA is both structurally principled and dynamically adaptive.\\
{\footnotesize Clarification: “Semantic genes” denote stable semantic substructures, not literal biological genes.}

\section{Challenges and Future Work}
Despite the strong efficiency gains observed in our experiments, several challenges remain and our study comes with important limitations:
\begin{itemize}
    \item \textbf{Dynamic Expert Management:} Future work must develop mechanisms for Codons (experts) to \textbf{dynamically create, prune, or merge} in real-time, mirroring biological adaptation.
    \item \textbf{End-to-End Differentiable Framework:} The ultimate goal is to integrate the energy cost ($L_c$) and SSI directly into the loss function, enabling the model to \textbf{learn the optimal energetic pathways} during training.
    \item \textbf{Experimental Scope:} Our evaluation is a proof-of-concept on 200 BioASQ documents, a single hardware platform (NVIDIA T4), and a specific baseline (Switch Transformer). Extending the analysis to larger corpora, additional tasks, and diverse hardware is an important direction for future work.
    \item \textbf{Measurement Limitations:} Energy metrics rely on GPU power traces obtained via \texttt{pynvml} under fixed experimental conditions. More systematic profiling, including variance estimates and alternative measurement tools, is needed to fully characterize energy behavior.
    \item \textbf{Strategic Perspective: Integration into SGEMAS (Future Work):} The efficiency of StructuredDNA is intended as a foundation for our next-generation architecture, the \textbf{Self-Growing Ephemeral Multi-Agent Systems (SGEMAS)}, a conceptual follow-up currently under development. The \textbf{98.8\%} EUD gain is expected to be essential for SGEMAS to guide its \textbf{Genesis Agent} in dynamically optimizing the \textbf{Time-to-Live (TTL)} of its ephemeral agents on devices like \textbf{NVIDIA Jetson}.
\end{itemize}

\section{Conclusion}
StructuredDNA illustrates how semantic routing can be derived from a bio-physical principle: inputs self-assemble into \emph{semantic codons} and computation is routed by minimizing a total energy functional. In our experiments, this mechanism yields stable, interpretable structures and selective activation across experts.

Empirically, StructuredDNA reduces per-token energy by \textbf{98.8\%} (EUD: $0.072351\!\rightarrow\!0.000835$ J/token) and inference time by \textbf{98.9\%} while maintaining high semantic stability (SSI $=\,\mathbf{0.998}$). Per-document energy drops from $\approx\,1332.745$ mJ to $\approx\,15.388$ mJ, despite a modest increase in instantaneous power (+2.0\%). Within the scope of our proof-of-concept, these results support energy minimization as a promising objective for modular, sparse AI, and suggest that in practice the optimal energy state coincides with an optimal semantic state.

Beyond energy, codon-level structure improves interpretability and opens the door to efficient on-device deployment. Our generalization experiments on WikiText-103 provide initial evidence for a \emph{Semantic Scaling Law}: in high-entropy domains, stability improves as expert granularity increases, while energy remains low. Future work will make $\tau$ and codon formation differentiable, integrate $L_c$ and SSI into the training loss, and enable dynamic expert creation/merging—extending the approach to larger models, broader domains, and real-time agentic systems, including massively sparse Transformers that can rival dense LLMs at a fraction of the energy cost.

The StructuredDNA implementation is publicly available at \url{https://github.com/InnoDeep-repos/StructuredDNA}, facilitating reproducibility and further research on energy-aware Transformer routing.


\end{document}